\documentclass{article}
\usepackage{look_read_enrich}

\usepackage{times}
\usepackage{soul}
\usepackage{url}
\usepackage[utf8]{inputenc}
\usepackage[small]{caption}
\usepackage{graphicx}
\usepackage{amsmath}
\usepackage{booktabs}
\usepackage{amssymb}
\usepackage{multirow}
\usepackage{subcaption}
\title{Look, Read and Enrich  \\
Learning from Scientific Figures and their Captions}

\author{
Jose Manuel Gomez-Perez \and
Raul Ortega \\
\affiliations
Expert System Cogito Labs\\
\emails
\{jmgomez,rortega\}@expertsystem.com
}

\begin{document}

\maketitle

\begin{abstract}
Compared to natural images, understanding scientific figures is particularly hard for machines. However, there is a valuable source of information in scientific literature that until now has remained untapped: the correspondence between a figure and its caption. In this paper we investigate what can be learnt by looking at a large number of figures and reading their captions, and introduce a figure-caption correspondence learning task that makes use of our observations. Training visual and language networks without supervision other than pairs of unconstrained figures and captions is shown to successfully solve this task. We also show that transferring lexical and semantic knowledge from a knowledge graph significantly enriches the resulting features. Finally, we demonstrate the positive impact of such features in other tasks involving scientific text and figures, like multi-modal classification and machine comprehension for question answering, outperforming supervised baselines and ad-hoc approaches. 
\end{abstract}
\section{Introduction}
\label{sec:intro}



Scientific knowledge is heterogeneous and can present itself in many forms, including text, mathematical equations, figures and tables. Like  many other manifestations of human thought, the scientific discourse usually adopts the form of a narrative, a scientific publication where related knowledge is presented in mutually supportive ways over different modalities. In the case of scientific figures, like charts, images and diagrams, these are usually accompanied by a text paragraph, a caption, that elaborates on the analysis otherwise visually represented. 


In this paper, we make use of this observation and tap on the potential of learning from the enormous source of free supervision available in the scientific literature, with millions of figures and their captions. We build models that learn from the scientific discourse both visually and textually by simply looking at the figures and reading their explanatory captions, inspired in how humans learn by reading a scientific publication. To this purpose, we explore how multi-modal scientific knowledge can be learnt from the correspondence between figures and captions. 

The main contributions of this paper are the following: 
\begin{itemize}
    \item \textbf{An unsupervised Figure-Caption Correspondence task (FCC)} that jointly learns text and visual features useful to address a range of tasks involving scientific text and figures. 
    \item \textbf{A method to enrich such features with semantic knowledge} transferred from structured knowledge graphs (KG).
    \item \textbf{A study of the complexity of figure-caption correspondence} compared to classical image-sentence matching.
    \item \textbf{A qualitative and quantitative analysis} of the learnt text and visual features through transfer learning tasks.
    \item \textbf{A corpus of scientific figures and captions} extracted from SN SciGraph and AI2 Semantic Scholar.
\end{itemize}


We present the FCC task in section~\ref{sec:alg}, including the network architecture, training protocol, and how adding pre-trained word and semantic embeddings can enrich the resulting text and visual features. In section~\ref{sec:res}, we first introduce our datasets and evaluate the performance of our method in the task it was trained to solve, the correspondence between scientific figures and captions. Then, we relate our work to the state of the art in image-sentence matching and evaluate our approach in two challenging transfer learning tasks: caption and figure classification and multi-modal machine comprehension. In section~\ref{sec:qual} we perform a qualitative study that illustrates how the FCC task leads to detailed textual and visual discrimination. Finally, in section~\ref{sec:con} we conclude the paper and advance future work.

\section{Related work}
\label{sec:rel}
Understanding natural images has been a major area of research in computer vision, with well established datasets like ImageNet~\cite{imagenet_cvpr09}, Flickr8K~\cite{Hodosh2013FramingID}, Flickr30K~\cite{Young2014FromID} and COCO~\cite{Lin2014MicrosoftCC}. However, reasoning with other visual representations like scientific figures and diagrams has not received the same attention yet and entails additional challenges: Scientific figures are more abstract and symbolic, their captions tend to be significantly longer and use specialized lexicon, and the relation between a scientific figure and its caption is unique, i.e. in a scientific publication there is only one caption that corresponds with one figure and vice versa.

The FCC task presented herein is a form of co-training~\cite{Blum1998CombiningLA} where there are two views of the data and each view provides complementary information. Similar two-branch neural architectures focus on image-sentence~\cite{Wang2018LearningTN,Eisenschtat2017} and audio-video~\cite{Arandjelovic2017LookLA} matching. Others like~\cite{Socher2013ZeroShotLT} learn common embeddings from images and text. However, in such cases one or both networks are typically pre-trained.


Focused on geometry,~\cite{Seo2014DiagramUI} maximize the agreement between text and visual data. In~\cite{Burns2018TowardsEE}, the authors apply machine vision and natural language processing to extract data from figures and their associated text in bio-curation tasks. In~\cite{Kembhavi2016ADI}, they parse diagram components and connectors as a Diagram Parse Graph (DPG), semantically interpret the DPG and use the model to answer diagram questions. While we rely on the correspondence between figures and captions, they train a specific classifier for each component and connector type and yet another model to ground the semantics of the DPG in each domain, like food webs or water cycles. 

Knowledge fusion approaches like~\cite{Thoma2017TowardsHC} investigate the potential of complementing KG embeddings with text and natural images by integrating information across the three modalities in a single latent representation. They assume pre-trained entity representations exist in each individual modality, e.g. the visual features encoding the image of a ball, the word embeddings associated to the token "ball", and the KG embeddings related to the ball entity, which are then stitched together. In contrast, FCC co-trains text and visual features from figures and their captions and supports the enrichment of such features with lexical and semantic knowledge transferred from a KG during the training of the FCC task. 

\section{Figure-Caption Correspondence}
\label{sec:alg}
The main idea of our approach is to learn a correspondence task between scientific figures and their captions as they appear in a scientific publication. The information captured in the caption explains the corresponding figure in natural language, providing guidance to identify the key features of the figure and vice versa. By seeing a figure and reading the textual description in its caption we ultimately aim to learn representations that capture e.g. what it means that two plots are similar or what gravity looks like. 

We leverage this observation to learn a figure-caption correspondence task. In essence, FCC is a binary classification task that receives a figure and a caption and determines whether they correspond or not. For training, the positive pairs are actual figures and their captions from a collection of scientific publications. Negative pairs are extracted from combinations of figures and any other randomly selected captions. The network is then made to learn text and visual features from scratch, without additional labelled data.


\subsection{FCC Architecture and Model}
\label{subsec:model}
\begin{figure}[t]
    \centering
    \includegraphics[width=0.5\textwidth]{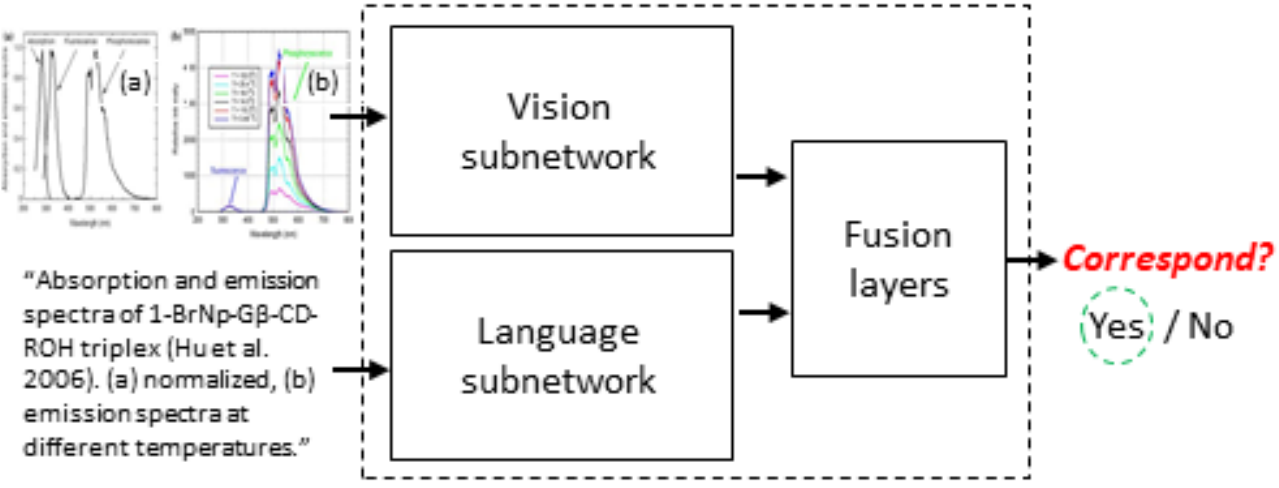}
    \caption{Proposed 2-branch architecture of the FCC task.}
   \label{fig:arch}
\end{figure}

We propose a 2-branch neural architecture (figure~\ref{fig:arch}) that has three main parts: the vision and language subnetworks, respectively extracting visual and text features, and a fusion subnetwork that takes the resulting features from the visual and text blocks and uses them to evaluate figure-caption correspondence. 

The \textbf{vision subnetwork} follows a VGG-style~\cite{Simonyan2014VeryDC} design, with 3x3 convolutional filters, 2x2 max-pooling layers with stride 2 and no padding. It contains 4 blocks of conv+conv+pool layers, where inside each block the two convolutional layers have the same number of filters, while consecutive blocks have doubling number of filters (64, 128, 256, 512). The input layer receives 224x224x3 images. The final layer produces a 512-D vector after 28x28 max-pooling. Each convolutional layer is followed by batch normalization~\cite{Ioffe2015BatchNA} and ReLU layers. Based on~\cite{Kim2014ConvolutionalNN}, the \textbf{language subnetwork} has 3 convolutional blocks, each with 512 filters and a 5-element window size with ReLU activation. Each convolutional layer is followed by a 5-max pooling layer, except for the final layer, which produces a 512-D vector after 35-max pooling. The language subnetwork has a 300-D embeddings layer at the input, with a maximum sequence length of 1,000 tokens. The \textbf{fusion subnetwork} calculates the element-wise product of the 512-D visual and text feature vectors into a single vector $r$ to produce a 2-way classification output (correspond or not). It has two fully connected layers, with ReLU and an intermediate feature size of 128-D. The probability of each choice is the softmax of $r$, i.e. $\hat{y} = softmax(r) \in \mathbb{R}^{2}$. During training, we minimize the negative log probability of the correct choice.

This architecture enables the FCC task to learn visual and text features from scratch in a completely unsupervised manner, just by observing the correspondence of figures and captions. Next, we extend it to enable the transfer of additional pre-trained information. Here, we focus on adding pre-trained embeddings on the language branch, and then back-propagate to the visual features during FCC training. Adding pre-trained visual features is also possible and indeed we also evaluate its impact in the FCC task in section~\ref{subsec:FCC}.

Let $V$ be a vocabulary of words from a collection of documents $D$. Also, let $L$ be their lemmas, i.e. base forms without morphological or conjugational variations, and $C$ the concepts (or senses) in a KG. Each word $w_k$ in $V$, e.g. {\em made}, has one lemma $l_k$ ({\em make}) and may be linked to one or more concepts $c_k$ in $C$ ({\em create or produce something}). 



For each word $w_k$, the FCC task learns a d-D embedding $\vec{w}_k$, which can be combined with pre-trained word ($\vec{w'}_k$), lemma ($\vec{l}_k$) and concept ($\vec{c}_k$) embeddings to produce a single vector $\vec{t}_k$. If no pre-trained knowledge is transferred from an external source, then $\vec{t}_k=\vec{w}_k$. Note that we previously lemmatize and disambiguate $D$ against the KG in order to select the right pre-trained lemma and concept embeddings for each particular occurrence of $w_k$. Equation~\ref{eq:agg_w_rep2} shows the different combinations of learnt and pre-trained embeddings we consider: (a) learnt word embeddings only, (b) learnt and pre-trained word embeddings and (c) learnt word embeddings and pre-trained semantic embeddings, including both lemmas and concepts, in line with our recent findings presented in~\cite{Denaux2019Vecsigrafo}.

\begin{equation}
\label{eq:agg_w_rep2}
\vec{t}_k =
\begin{cases}
    \vec{w}_k & \quad (a)\\
    [\vec{w}_k; \vec{w'}_k] & \quad (b)\\
    [\vec{w}_k; \vec{l}_k; \vec{c}_k] & \quad (c)\\
\end{cases}
\end{equation}

In our experiments, concatenation proved optimal to combine the embeddings learnt by the network and the pre-trained embeddings, compared to other methods like summation, multiplication, average or learning a task-specific weighting of the different representations as in~\cite{Peters}. Since some words may not have associated pre-trained word, lemma or concept embeddings, we pad these sequences with $\varnothing_W$, $\varnothing_L$ and $\varnothing_C$, which are never included in the vocabulary. The dimensionality of $\vec{t}_k$ is fixed to 300, i.e. the size of each sub-vector in configurations $(a)$, $(b)$ and $(c)$ is 300, 150 and 100, respectively. In doing so, we aimed at limiting the number of trainable parameters and balance the contribution of each information source.  

In its most basic form, i.e. configuration $(a)$, the FCC network has over 32M trainable parameters (28M in the language subnetwork, 4M in the vision subnetwork and 135K in the fusion subnetwork) and takes 12 hours to train on a single GPU Nvidia GeForce RTX 2080 Ti for a relatively small corpus (SN SciGraph, see section~\ref{subsec:data}). We used 10-fold cross validation, Adam optimization~\cite{Kingma2014AdamAM} with learning rate $10^{-4}$ and weight decay $10^{-5}$. The network was implemented\footnote{All the code and data, including the corpora extracted from SciGraph and Semantic Scholar, are available through \url{https://github.com/hybridNLP/look_read_and_enrich}} in Keras and TensorFlow, with batch size 32. The number of positive and negative cases is balanced within the batches. 

\subsection{Semantic Embeddings}
\label{subsec:kg}
We use HolE~\cite{Nickel2016HolographicEO} and Vecsigrafo~\cite{Denaux2019Vecsigrafo} to learn semantic embeddings. The latter extends the Swivel algorithm~\cite{Shazeer2016SwivelIE} to jointly learn word, lemma and concept embeddings on a corpus disambiguated against the KG, outperforming the previous state of the art in word and word-sense embeddings by co-training word, lemma and concept embeddings as opposed to training each individually. In contrast to Vecsigrafo, which requires both a text corpus and a KG, HolE follows a graph-based approach where embeddings are learnt exclusively from the KG. As section~\ref{subsec:FCC} will show, this gives Vecsigrafo a certain advantage in the FCC task. Following up with the work presented in~\cite{Denaux2019Vecsigrafo}, our experiments focus on Sensigrafo, the KG underlying Expert System's Cogito NLP proprietary platform. Similar to WordNet, on which Vecsigrafo has also been successfully trained\footnote{See our tutorial on hybrid NLP at \url{http://expertsystemlab.com/hybridNLP18}}, Sensigrafo is a general-purpose KG with lexical and semantic information that contains over 300K concepts, 400K lemmas and 80 types of relations rendering 3M links. We use Cogito to disambiguate the text corpora prior to training Vecsigrafo. All the semantic (lemma and concept) embeddings produced with HolE or Vecsigrafo are 100-D.

\section{Results and Discussion}
\label{sec:res}
In this section, first we evaluate the actual FCC task against two supervised baselines. Then, we situate our work in the more general image-sentence matching problem, showing empirical evidence of the additional complexity associated to the scientific domain and the figure-caption case compared to natural images. Next, we test the visual and text features learnt in the FCC task in two different transfer learning settings: classification of scientific figures and captions and multi-modal machine comprehension for question answering given a context of text, figures and images. 

\subsection{Datasets} 
\label{subsec:data}
We have used the following datasets for training and evaluation:

\textbf{The Semantic Scholar corpus}~\cite{Ammar2018ConstructionOT} (SemScholar) is a large dataset of scientific publications made available by AI2. From its 39M articles, we downloaded 3,3M PDFs (the rest were behind paywalls, did not have a link or it was broken) and extracted 12.5M figures and captions through PDFFigures2~\cite{Clark2016PDFFigures2M}. We randomly selected 500K papers to train the FCC task on their figures and captions and another 500K to train Vecsigrafo on the text of their titles and abstracts.

\textbf{Springer Nature's SciGraph}\footnote{\url{https://www.springernature.com/gp/researchers/scigraph}} contains 7M scientific publications organized in 22 scientific fields or categories. Since SciGraph does not provide a link to the PDF of the publication, we selected the intersection with SemScholar, producing a smaller corpus of 80K papers (in addition to the 1M papers from SemScholar mentioned above) and 82K figures that we used for training certain FCC configurations and supervised baselines (section~\ref{subsec:FCC}).

\textbf{The Textbook Question Answering corpus}~\cite{Kembhavi2017AreYS} includes 1,076 lessons and 26,260 multi-modal test questions from middle school science curricula. Its complexity and scope  make it a challenging textual and visual question answering dataset. 

\textbf{Wikipedia.} We used the January 2018 English Wikipedia dataset as one of the corpora on which to train Vecsigrafo. As opposed to SciGraph or SemScholar, specific of the scientific domain, Wikipedia is a source of general-purpose information. 

\textbf{Flickr30K and COCO}, as image-sentence matching benchmarks.

\subsection{Figure-Caption Correspondence}
\label{subsec:FCC}
We evaluate our method in the task it was trained to solve: determining whether a figure and a caption correspond. We also compare the performance of the FCC task against two supervised baselines, training them on a classification task against the SciGraph taxonomy. For such baselines we first train the vision and language networks independently and then combine them. The feature extraction parts of both networks are the same as described in section~\ref{subsec:model}. On top of them, we attach a fully connected layer with 128 neurons and ReLU activation and a softmax layer, with as many neurons as target classes. 

The \textbf{direct combination} baseline computes the figure-caption correspondence through the scalar product between the softmax outputs of both networks. If it exceeds a threshold, which we heuristically fixed on 0.325, the result is positive. The \textbf{supervised pre-training} baseline freezes the weights of the feature extraction trunks from the two trained networks, assembles them in the FCC architecture as shown in section~\ref{subsec:model}, and trains the FCC task on the fully connected layers. While direct combination provides a notion of the agreement between the two branches, supervised pre-training is the most similar supervised approach to our method.

\begin{table}[]
\small
\centering
\begin{tabular}{@{}lcccc@{}}
\toprule
 & Corpus & Word rep. & Acc$_{vgg}$. & Acc. \\ \midrule
Direct & \multirow{2}{*}{SciGraph} & $\vec{w}_k$ & \multicolumn{2}{c}{60.30}\\
Pre-train & & $\vec{w}_k$ & \multicolumn{2}{c}{68.40} \\ \midrule
$FCC_1$ & \multirow{4}{*}{SciGraph} & $\vec{w}_k$ & 78.09 & 78.48 \\
$FCC_2$ &  & $[\vec{w}_k; \vec{w'}_{k\_sem}]$ & 79.75 & 80.35\\
$FCC_3$ &  & $[\vec{w}_k; \vec{l}_{k\_holE}; \vec{c}_{k\_holE}]$ & 78.64 & 78.08\\
$FCC_4$ &  & $[\vec{w}_k; \vec{l}_{k\_wiki}; \vec{c}_{k\_wiki}]$ & 79.71 & 80.50 \\
$FCC_5$ &  & $[\vec{w}_k; \vec{l}_{k\_sem}; \vec{c}_{k\_sem}]$ & \textbf{80.50} & \textbf{81.97}\\ \midrule
$FCC_6$ & \multirow{2}{*}{SemScholar} & $\vec{w}_k$ & 80.42 & 81.44\\
$FCC_7$ & & $[\vec{w}_k; \vec{l}_{k\_sem}; \vec{c}_{k\_sem}]$ & \textbf{82.21} & \underline{\textbf{84.34}} \\
\bottomrule
\end{tabular}
\caption{FCC and supervised baselines results (\% accuracy).}
\label{table:corr}
\end{table}

Table~\ref{table:corr} shows the results of the FCC task and the supervised baselines. $FCC_k$ denotes the corpus and word representation used to train the FCC task. Acc$_{vgg}$ shows the accuracy after replacing our visual branch with pre-trained VGG16 features learnt on ImageNet. This provides an estimate of how specific of the scientific domain scientific figures and therefore the resulting visual features can be, compared to natural images. As the table shows, the results obtained using pre-trained visual features are clearly worse in general (only slightly better in $FCC_3$), suggesting that the visual information contained in scientific figures indeed differs from natural images.

We trained the FCC network on two different scientific corpora: SciGraph ($FCC_{1-5}$) and SemScholar ($FCC_{6-7}$). Both $FCC_1$ and $FCC_6$ learnt their own word representations without transfer of any pre-trained knowledge. Even in its most basic form our approach substantially improves over the supervised baselines, confirming that the visual and language branches learn from each other and also that figure-caption correspondence is an effective source of free supervision.


Adding pre-trained knowledge at the input layer of the language subnetwork provides an additional boost, particularly with lemma and concept embeddings from Vecsigrafo ($FCC_5$). Vecsigrafo clearly outperformed HolE ($FCC_3$), which was also beaten by pre-trained fastText~\cite{Bojanowski2017EnrichingWV} word embeddings ($FCC_2$) trained on SemScholar. 

Since graph-based KG embedding approaches like HolE only generate embeddings of the artifacts explicitly contained in the KG, this may indicate that Sensigrafo, the KG used in this task, provides a partial coverage of the scientific domain, as could be expected since we are using an off-the-shelf version. Deeper inspection shows that HolE only covers 20\% of the lemmas in the SciGraph vocabulary. On the other hand, Vecsigrafo, trained on the same KG, also captures lexical information from the text corpora it is trained on, Wikipedia or SemScholar, raising lemma coverage to 42\% and 47\%, respectively.

Although the size of Wikipedia is almost triple of our SemScholar corpus, training Vecsigrafo on the latter resulted in better FCC accuracy  ($FCC_4$ vs. $FCC_5$), suggesting that domain relevance is more significant than sheer volume, in line with our previous findings in~\cite{DBLP:conf/esws/Garcia-SilvaG18}. Training FCC on SemScholar, much larger than SciGraph, further improves accuracy, as shown in $FCC_6$ and $FCC_7$.



\subsection{Image-Sentence Matching}
\label{subsec:img-sent}
We put our FCC task in the context of the more general problem of image-sentence matching through a bidirectional retrieval task where images are sought given a text query and vice versa. While table~\ref{table:bresnid} focuses on natural images datasets (Flickr30K and COCO), table~\ref{table:bressd} shows results on scientific datasets (SciGraph and SemScholar) rich in scientific figures and diagrams. The selected baselines (Embedding network, 2WayNet, VSE++ and DSVE-loc) report results obtained on the Flickr30K and COCO datasets, which we also include in table~\ref{table:bresnid}. Performance is measured in recall at k ($Rk$), with k=\{1,5,10\}. From the baselines, we successfully reproduced DSVE-loc, using the code made available by the authors\footnote{\url{https://github.com/technicolor-research/dsve-loc}}, and trained it on SciGraph and SemScholar. 

We trained the FCC task on all the datasets, both in a totally unsupervised way and with pre-trained semantic embeddings\footnote{For conciseness, we focus on the best FCC configuration in table~\ref{table:corr}, based on Vecsigrafo.} (indicated with subscript $vec$), and executed the bidirectional retrieval task using the resulting text and visual features. We also experimented with pre-trained VGG16 visual features extracted from ImageNet (subscript $vgg$), with more than 14 million hand-annotated images. Following common practice in image-sentence matching, our splits are 1,000 samples for test and the rest for training.

We can see a marked division between the results obtained on natural images datasets (table~\ref{table:bresnid}) and those focused on scientific figures (table~\ref{table:bressd}). In the former case, VSE++ and DSVE-loc clearly beat all the other approaches. In contrast, our model performs poorly on such datasets although results are ameliorated when we use pre-trained visual features from ImageNet ("Ours\textsubscript{vgg}" and "Ours\textsubscript{vgg-vec}"). Interestingly, the situation reverts with the scientific datasets. While the recall of DSVE-loc drops dramatically in SciGraph, and even more in SemScholar, our approach shows the opposite behavior in both figure and caption retrieval. Using visual features enriched with pre-trained semantic embeddings from Vecsigrafo during training of the FCC task further improves recall in the bidirectional retrieval task. Compared to natural images, the additional complexity of scientific figures and their caption texts, which in addition are considerably longer (see table~\ref{table:caplen}), seems to have a clear impact in this regard.

\begin{table}[h]
\small
\centering
\begin{tabular}{@{}lcccc@{}}
\toprule
& Flickr30K & COCO & SciGraph & SemScholar \\ \cmidrule(lr){2-3} \cmidrule(lr){4-5}
Max & 81 & 179 & 514 & 836 \\
Mean & 12 & 11 & 28 & 42 \\
\bottomrule
\end{tabular}
\caption{Caption length: natural images vs scientific datasets.
}
\label{table:caplen}
\end{table}

Unlike in Flickr30K and COCO, replacing the FCC visual features with pre-trained ones from ImageNet brings us little benefit in SciGraph and even less in SemScholar, where the combination of FCC and Vecsigrafo ("Ours\textsubscript{vec}") obtains the best results across the board. This and the extremely poor performance of the best image-sentence matching baseline (DSVE-loc) in the scientific datasets shows evidence that dealing with scientific figures is considerably more complex than natural images. Indeed, the best results in figure-caption correspondence ("Ours\textsubscript{vec}" in SemScholar) are still far from the SoA in image-sentence matching (DSVE-loc in COCO).

\begin{table*}[h]
\footnotesize
\centering
\begin{tabular}{lcccccccccccc}
\toprule
 \multirow{3}{*}{\footnotesize{Model}} & \multicolumn{6}{c}{\footnotesize{Flickr30K}} & \multicolumn{6}{c}{\footnotesize{COCO}} \\ \cmidrule(lr){2-7} \cmidrule(lr){8-13} 
  & \multicolumn{3}{c}{Caption-to-image} & \multicolumn{3}{c}{Image-to-caption} & \multicolumn{3}{c}{Caption-to-image} & \multicolumn{3}{c}{Image-to-caption} \\ \cmidrule(lr){2-4} \cmidrule(lr){5-7} \cmidrule(lr){8-10} \cmidrule(lr){11-13}
  & R1 & R5 & R10 & R1 & R5 & R10 & R1 & R5 & R10 & R1 & R5 & R10 \\ \cmidrule(lr){1-13}
 Emb. net~\cite{Wang2018LearningTN} & 29.2 & 59.6 & 71.7 & 40.7 & 69.7& 79.2 & 39.8 & 75.3 & 86.6 & 50.4 & 79.3 & 89.4 \\
 2WayNet~\cite{Eisenschtat2017} & 36.0 & 55.6 & n/a & 49.8 & 67.5 & n/a & 39.7 & 63.3 & n/a & 55.8 & 75.2 & n/a \\
 VSE++~\cite{Faghri2017VSEIV} & \textbf{39.6} & n/a & \textbf{79.5} & \textbf{52.9} & n/a & \textbf{87.2} & 52.0 & n/a & 92.0 & 64.6 & n/a & 95.7 \\ \cmidrule(lr){1-13}
 DSVE-loc~\cite{Engilberge2018FindingBI} & 34.9 & \textbf{62.4} & 73.5 & 46.5 & \textbf{72.0} & 82.2 & \textbf{55.9} & \textbf{86.9} & \textbf{94} & \textbf{69.8} & \textbf{91.9} & \textbf{96.6} \\ \cmidrule(lr){1-13}
 Ours\textsubscript{vgg} & 3.4 & 14.0 & 23.2 & 4.7 & 16.4 & 24.8 & 11.7 & 39.7 & 58.8 & 15.2 & 40.0 & 56.1 \\ 
 Ours & 0.4 & 1.3 & 2.8 & 0.2 & 1.5 & 3.2 & 2.6 & 10.3 & 18.0 & 2.5 & 9.3 & 17.3 \\
 Ours\textsubscript{vgg-vec} & 5.4 & 17.8 & 27.8 & 6.8 & 20.3 & 32.0 & 12.8 & 40.9 & 59.7 & 17.3 & 41.2 & 57.4 \\
 Ours\textsubscript{vec} & 0.6 & 2.9 & 5.3 & 1.2 & 3.7 & 6.5 & 4.0 & 14.6 & 25.3 & 4.4 & 15.6 & 25.9 \\ 
 \bottomrule
\end{tabular}
\caption{Bidirectional retrieval. FCC vs. image-sentence matching baselines (\%recall$@$k). Natural images datasets.}
\label{table:bresnid}
\end{table*}

\begin{table*}[h]
\footnotesize
\centering
\begin{tabular}{lcccccccccccc}
\toprule
 \multirow{3}{*}{\footnotesize{Model}} & \multicolumn{6}{c}{\footnotesize{SciGraph}} & \multicolumn{6}{c}{\footnotesize{SemScholar}} \\ \cmidrule(lr){2-7} \cmidrule(lr){8-13} 
  & \multicolumn{3}{c}{Caption-to-figure} & \multicolumn{3}{c}{Figure-to-caption} & \multicolumn{3}{c}{Caption-to-figure} & \multicolumn{3}{c}{Figure-to-caption} \\ \cmidrule(lr){2-4} \cmidrule(lr){5-7} \cmidrule(lr){8-10} \cmidrule(lr){11-13}
  & R1 & R5 & R10 & R1 & R5 & R10 & R1 & R5 & R10 & R1 & R5 & R10 \\ \cmidrule(lr){1-13}
 Emb. net~\cite{Wang2018LearningTN} & n/a & n/a & n/a & n/a & n/a & n/a & n/a & n/a & n/a & n/a & n/a & n/a \\
 2WayNet~\cite{Eisenschtat2017} & n/a & n/a & n/a & n/a & n/a & n/a & n/a & n/a & n/a & n/a & n/a & n/a  \\
 VSE++~\cite{Faghri2017VSEIV} & n/a & n/a & n/a & n/a & n/a & n/a & n/a & n/a & n/a & n/a & n/a & n/a   \\ \cmidrule(lr){1-13}
 DSVE-loc~\cite{Engilberge2018FindingBI} & 0.7 & 3.1 & 5.3 & 1.4 & 1.4 & 2.4 & 0.9 & 3 & 4.5 & 0.8 & 0.8 & 1.3\\ \cmidrule(lr){1-13}
 Ours\textsubscript{vgg} & 1.4 & 6.6 & 11.3 & 1.3 & 6.4 & 10.6 & 2.9 & 9.5 & 17.4 & 3.1 & 12.1 & 18.0 \\ 
 Ours & 0.7 & 5.7 & 11.4 & 1.2 & 4.9 & 10.0 & 2.8 & 11.4 & 18.8 & 2.1 & 10.6 & 18.2\\
 Ours\textsubscript{vgg-vec} & \textbf{1.7} & 7.8 & 14.2 & 2.1 & 7.7 & 15.8 & 2.9 & 13.9 & 24.0 & \textbf{4.7} & 14.9 & 23.2 \\
 Ours\textsubscript{vec} & 1.5 & \textbf{9.0} & \textbf{14.9} & \textbf{2.6} & \textbf{9.7} & \textbf{16.1} & \textbf{3.9} & \textbf{15.5} & \textbf{25.1} & 4.4 & \textbf{16.6} & \textbf{25.6} \\ 
 \bottomrule
\end{tabular}
\caption{Bidirectional retrieval. FCC vs. image-sentence matching baselines (\%recall$@$k). Scientific datasets.}
\label{table:bressd}
\end{table*}

\subsection{Caption and Figure Classification} 
\label{subsec:class}
We evaluate the language and visual representations emerging from FCC in the context of two classification tasks that aim to identify the scientific field an arbitrary text fragment (a caption) or a figure belong to, according to the SciGraph taxonomy. The latter is a particularly hard task due to the whimsical nature of the figures that appear in our corpus: figure and diagram layout is arbitrary; charts, e.g. bar and pie charts, are used to showcase data in any field from health to engineering; figures and natural images appear indistinctly, etc. Also, note that we only rely on the actual figure, not the text fragment  where it is mentioned in the paper.

We pick the text and visual features that produced the best FCC results with and without pre-trained semantic embeddings (table~\ref{table:corr}, $FCC_7$ and $FCC_6$, respectively) and use the language and vision subnetworks presented in section~\ref{subsec:model} to train our classifiers on SciGraph in two different scenarios. First, we only fine tune the fully connected and softmax layers, freezing the text and visual weights (non-trainable in the table). Second, we fine tune all the parameters in both networks (trainable). In both cases, we compare against a baseline using the same networks initialized with random weights, without FCC training. In doing so, through the first, non-trainable scenario, we seek to quantify the information contributed by the FCC features, while training from scratch on the target corpus should provide an upper bound for figure and caption classification. Additionally, for figure classification, we select a baseline of frozen VGG16 weights trained on ImageNet. We train using 10-fold cross validation and Adam. For the caption classification task, we select learning rate $10^{-3}$ and batch size 128. In figure classification, we use learning rate $10^{-4}$, weight decay $10^{-5}$ and batch size 32.

\begin{table}[h]
\small
\centering
\begin{tabular}{@{}lcccc@{}}
\toprule
\multirow{2}{*}{Model} & \multicolumn{2}{c}{Caption} & \multicolumn{2}{c}{Figure}\\ \cmidrule(lr){2-3} \cmidrule(lr){4-5}
 & Non-trainable & Trainable & Non-trainable & Trainable \\ \midrule
 Random & 39.92 & 78.20 & 44.19 & 61.21 \\
 VGG16 & n/a & n/a & 58.43 & n/a \\ \midrule
 Ours FCC6 & 61.31 & \textbf{\underline{79.24}} & 58.57 & \textbf{\underline{63.60}} \\
 Ours FCC7 & \textbf{67.40} & 79.11 & \textbf{60.19} & 63.49 \\ \bottomrule
\end{tabular}
\caption{Caption and figure classification (\%accuracy)}
\label{table:textlangclass}
\end{table}

The results in table~\ref{table:textlangclass} show that our approach amply beats the baselines, including the upper bound (training from scratch on SciGraph). The delta is particularly noticeable in the non trainable case for both caption and figure classification and is considerably increased in "Ours $FCC_7$", which uses pre-trained semantic embeddings. This includes both the random and VGG baselines and illustrates again the additional complexity of analyzing scientific figures compared to natural images, even if the latter is trained on a considerably larger corpus like ImageNet. Fine tuning the whole networks on SciGraph further improves accuracies. In this case, "Ours $FCC_6$", which uses FCC features without additional pre-trained embeddings, slightly outperforms "Ours $FCC_7$", suggesting a larger margin to learn from the task-specific corpus. Note that both $FCC_6$ and $FCC_7$ were trained on SemScholar.

\begin{table*}[t]
\footnotesize
\centering
\begin{tabular}{lccccccc}
\toprule
 Model & Text & Visual & Word representation & Figure rep. & Inspired by & MC$_{text}$ & MC$_{diag}$ \\ \midrule
Random & x & x & n/a & n/a & Random & 22.7 & 25.0\\ 
BiDAF & \checkmark & x & $\vec{w}_k$ & n/a & BiDAF~\cite{Seo2017BidirectionalAF} & 32.2 & 30.1\\ \midrule
$TQA_1$ & \checkmark & x & \multirow{5}{*}{$\vec{w}_k$} & n/a & MemoryNet~\cite{Weston2014MemoryN} & 32.9 & 29.9\\
$TQA_2$ & \checkmark & \checkmark & & VGG19 & VQA~\cite{Antol2015VQAVQ} & n/a & 29.9\\ 
$TQA_3$ & \checkmark & \checkmark &  & DPG & DSDP-NET~\cite{Kembhavi2017AreYS} & n/a & 31.3\\ 
$TQA_4$ & \checkmark & \checkmark &  & $FCC_6$  & \multirow{2}{*}{FCC} & 33.89 & 34.27 \\
$TQA_5$ & \checkmark & \checkmark &  & $FCC_7$ &  & 33.73 & \em 33.52 \\ \midrule
$TQA_6$ & \checkmark & x & \multirow{4}{*}{$[\vec{w}_k+\vec{l}_{k\_sem}+\vec{c}_{k\_sem}]$} & n/a & MemoryNet~\cite{Weston2014MemoryN} & 35.41 & 34.57 \\
$TQA_7$ & \checkmark & \checkmark & & VGG19 & VQA~\cite{Antol2015VQAVQ} & 36.26 & 32.58 \\ 
$TQA_8$ & \checkmark & \checkmark &  & DPG & DSDP-NET~\cite{Kembhavi2017AreYS} & n/a & n/a\\ 
$TQA_9$ & \checkmark & \checkmark & & $FCC_6$ & \multirow{2}{*}{FCC} & \textbf{36.56} & \textbf{35.30} \\
$TQA_{10}$ & \checkmark & \checkmark & & $FCC_7$ & & 35.84 & 33.94 \\ 
\bottomrule
\end{tabular}
\caption{TQA results (\% accuracy). FCC vs. random, BiDAF, MemoryNet, VQA and DSDP-NET baselines.}
\label{table:tqa}
\end{table*}

\subsection{Textbook Question Answering (TQA) for Multi-Modal Machine Comprehension} 
\label{subsec:TQA}

We leverage the TQA dataset and the baselines in~\cite{Kembhavi2017AreYS} to evaluate the features learnt by the FCC task in a multi-modal machine comprehension scenario. We study how our model, which was not originally trained for this task, performs against state of the art models specifically trained for diagram question answering and textual reading comprehension in a very challenging dataset. We also study how pre-trained semantic embeddings impact in the TQA task: first, by enriching the visual features learnt in the FCC task as shown in section~\ref{subsec:model} and then by using pre-trained semantic embeddings to enrich word representations in the TQA corpus. 

We focus on multiple-choice questions, 73\% of the dataset. Table~\ref{table:tqa} shows the performance of our model against the results reported in~\cite{Kembhavi2017AreYS} for five TQA baselines: random, BiDAF (focused on text machine comprehension), text only ($TQA_1$, based on MemoryNet), text+image ($TQA_2$, VQA), and text+diagrams ($TQA_3$, DSDP-NET). We successfully reproduced the $TQA_1$ and $TQA_2$ architectures and adapted the latter\footnote{While VGG19 produces a 7-by-7 grid of 512-D image patch vectors, our visual subnetwork produces a 512-D vector. To align dimensions, we add a 7-max pooling layer.}. Then, we replaced the visual features in $TQA_2$ with those learnt by the FCC visual subnetwork both in a completely unsupervised way ($FCC_6$ in table~\ref{table:corr}) and with pre-trained semantic embeddings ($FCC_7$), resulting in $TQA_4$ and $TQA_5$, respectively. 

While $TQA_{1-5}$ used no pre-trained embeddings at all, $TQA_{6-10}$ were trained including pre-trained Vecsigrafo semantic embeddings. Unlike FCC, where we used concatenation to combine pre-trained lemma and concept embeddings with the word embeddings learnt by the task, element-wise addition worked best in the case of TQA. 

Following the recommendations in~\cite{Kembhavi2017AreYS}, we pre-processed the TQA corpus to i) consider knowledge from previous lessons in the textbook in addition to the lesson of the question at hand and ii) address challenges like long question contexts with a large lexicon. In both text and diagram MC, applying the Pareto principle to reduce the maximum token sequence length in the text of each question, their answers and context improved accuracy considerably. This optimization allowed reducing the amount of text to consider for each question, improving the signal to noise ratio. Finally, we obtained the most relevant paragraphs for each question through tf-idf and trained the models using 10-fold cross validation, Adam, learning rate $10^{-2}$ and batch size 128. In text MC we also used 0.5 dropout and recurrent dropout in the LSTM layers. 

Fitting multi-modal sources into a single memory, the use of visual FCC features clearly outperforms all the TQA baselines in diagram MC. Enhancing word representation with pre-trained semantic embeddings during training of the TQA task provides an additional boost that results in the highest accuracies for both text MC and diagram MC. These are significantly good results since, according to the TQA authors~\cite{Kembhavi2017AreYS}, most diagram questions in the TQA corpus would normally require a specific rich diagram parse, which we did not aim to provide.

\section{Qualitative Analysis} 
\label{sec:qual}


\begin{figure*}[t]
\centering
   \includegraphics[width=\textwidth]{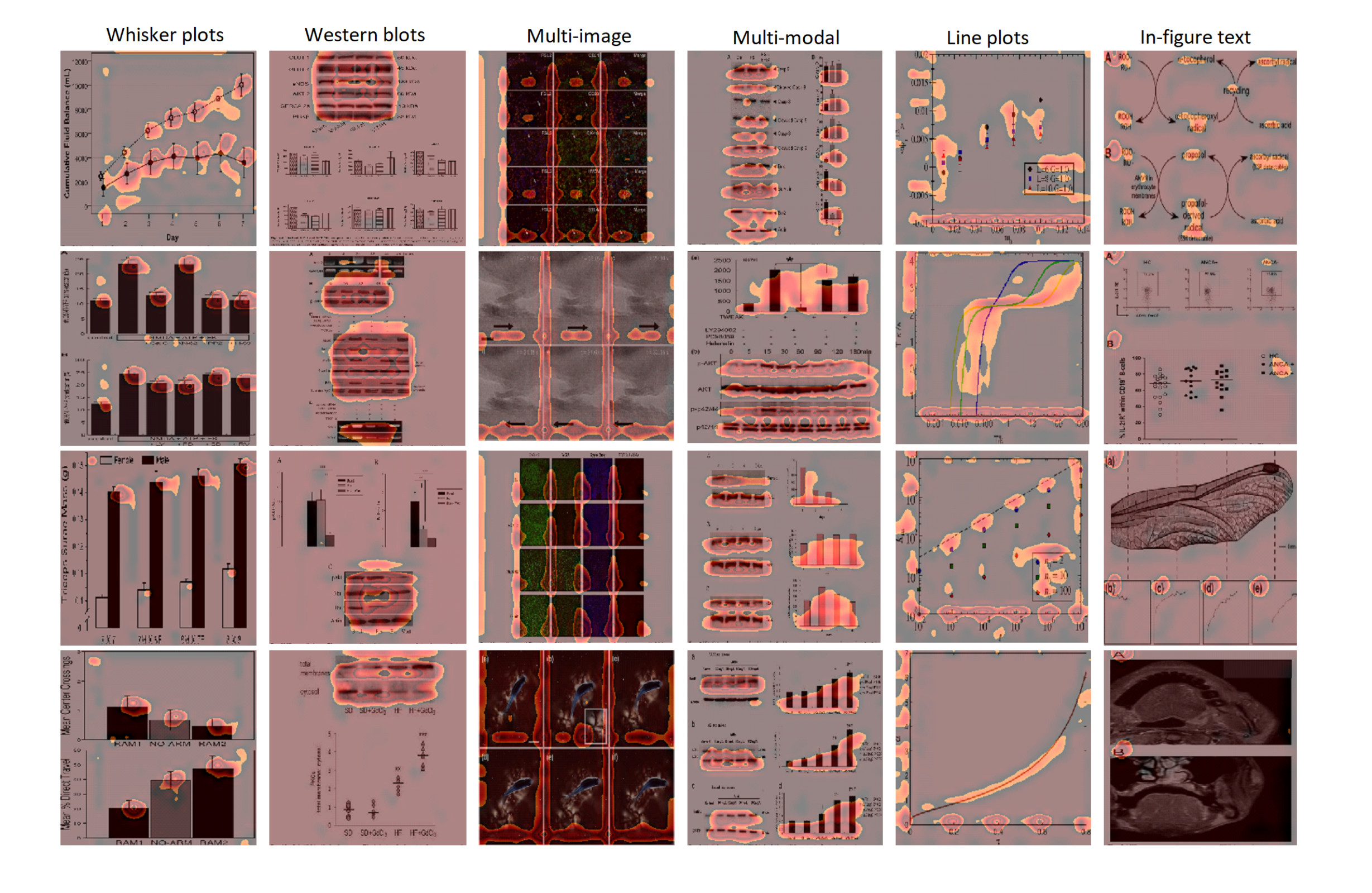}
   \vspace{-2.5mm}
   \caption{Selected visual features and activation heatmaps. The top row labels the dominant pattern for each feature.}
   \label{fig:featheat}
\end{figure*}

We inspect the features learnt by our FCC task to gain a deeper understanding of the syntactic and semantic patterns captured for figure and caption representation. The findings reported herein are qualitatively consistent for all the FCC variations in table~\ref{table:corr}.

\textbf{Vision features.} The analysis was carried out on an unconstrained variety of charts, diagrams and natural images from SciGraph, without filtering by figure type or scientific field. To obtain a representative sample of what the FCC network learns, we focus on the 512-D vector resulting from the last convolutional block before the fusion subnetwork. We pick the features with the most significant activation over the whole dataset and select the figures that activate them most. To this purpose, we prioritize those with higher maximum activation against the average activation.


Figure~\ref{fig:featheat} shows a selection of 6 visual features with the 4 figures that activate each feature more significantly and their activation heatmaps. Only figures are used as input, no text. As can be seen, the vision subnetwork has automatically learnt, without explicit supervision, to recognize different types of diagrams, charts and content, such as (from left to right) whisker plots, western blots (a technique used to identify proteins in a tissue sample), multi-image comparison diagrams, multi-modal data visualization charts (e.g. western plots vs. bar charts), line plots, and text within the figures. Furthermore, as shown by the heatmaps, our model discriminates the key elements associated to the figures that most activate each feature: the actual whiskers, the blots, the borders of each image under comparison, the blots and their complementary bar charts, as well as the line plots and the correspondence between them and the values in the x and y axes. Also, see (right-most column) how a feature discriminates text inserted in the figure, regardless of the remaining elements that may appear and the connections between them. This shows evidence of how the visual features learnt by the FCC task support the parsing of complex scientific diagrams. 

We also estimated a notion of semantic specificity based on the concepts of a KG. For each visual feature, we aggregated the captions of the figures that most activate it and used Cogito to disambiguate the Sensigrafo concepts that appear in them. Then, we estimated how important each concept is to each feature by calculating its tf-idf. Finally, we averaged the resulting values to obtain a consolidated semantic specificity score per feature. 

The scores of the features in figure~\ref{fig:featheat} range between 0.42 and 0.65, which is consistently higher than average (0.4). This seems to indicate a correlation between activation and the semantic specificity of each visual feature. For example, the heatmaps of the figures related to the feature with the lowest tf-idf (left-most column) highlights a particular visual pattern, i.e. the {\em whiskers}, that may spread over many, possibly unrelated domains. On the other hand, the feature with the highest score (second column) focuses on a type of diagrams, western blots, almost exclusive of protein and genetic studies. Others, like the feature illustrated by the figures in the fifth column, capture the semantics of a specific type of 2D charts relating two magnitudes {\em x} and {\em y}. Analyzing their captions with Cogito, we see that concepts like e.g. {\em isochronal} and {\em exponential functions} are mentioned. If we look at the second and four top-most figures in the column, we can see that such concepts are also visually depicted in the figures, suggesting that the FCC task has learnt to recognize them both from the text and visually. 

\begin{figure*}[t]
\centering
   \includegraphics[width=\textwidth]{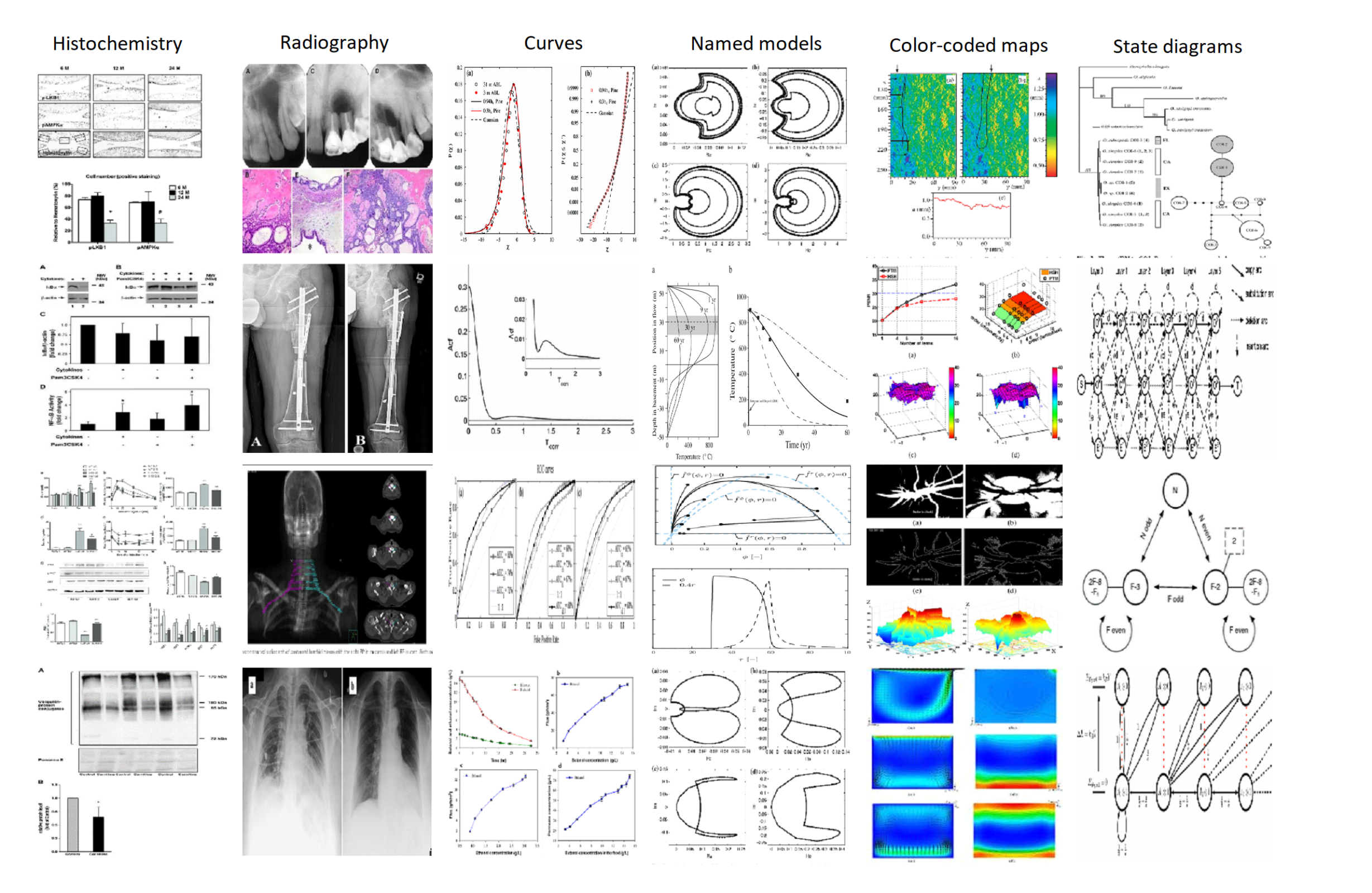}
   \caption{Selected text features. Top row labels the dominant pattern for each text feature.}
   \label{fig:textfeat}
\end{figure*}

\textbf{Text features.} Similar to the visual case, we selected the features from the last block of the language subnetwork with the highest activation. For visualization purposes, we picked the figures corresponding to the captions in SciGraph that most activate such features (figure~\ref{fig:textfeat}). No visual information is used. 

Several distinct patterns emerge from the text. The text feature in the first column seems to focus on genetics and histochemistry, including terms like {\em western blots} or {\em immunostaining} and variations like {\em immunoblot-s/ted/ting}. Interestingly, it also seems to have learnt some type of {\em is-a} relations (western blot is a type of immunoblot). The second feature focuses on variations of the term {\em radiograph}, e.g. {\em radiograph-y/s}. The third feature specializes in text related to curve plots involving several statistic analysis, e.g.  {\em Real-time PCR, one-way ANOVA} or {\em Gaussian distribution}. Sometimes (fourth figure from top) the caption does not mention the plot directly, but focuses on the analysis instead, e.g. "the data presented here are mean values of duplicate experiments", indicating transfer of knowledge from the visual part during training. The fourth feature extracts citations and models named after prominent scientists, e.g. {\em Evans function} (first and fourth figure), {\em Manley (1992)} (second), and {\em Aliev-Panfilov model} (third). The fifth feature extracts chromatography terminology, e.g. {\em 3D surface plot, photomicrograph} or {\em color map} and, finally, the right-most feature focuses on different types of named diagrams, like flow charts and state diagrams, e.g. {\em phylogenetic trees}. 

\begin{figure*}[h]
    \centering
    \includegraphics[width=\textwidth]{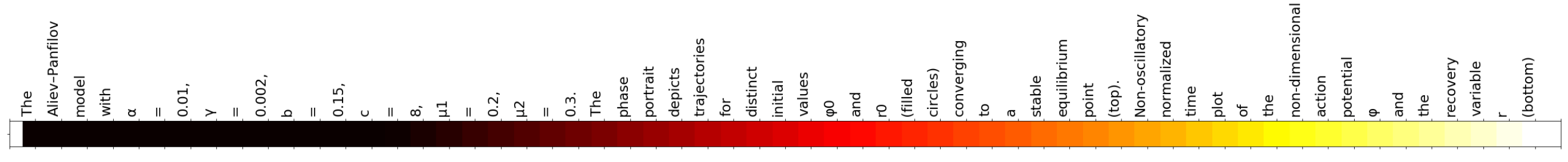}\\
    \includegraphics[width=\textwidth]{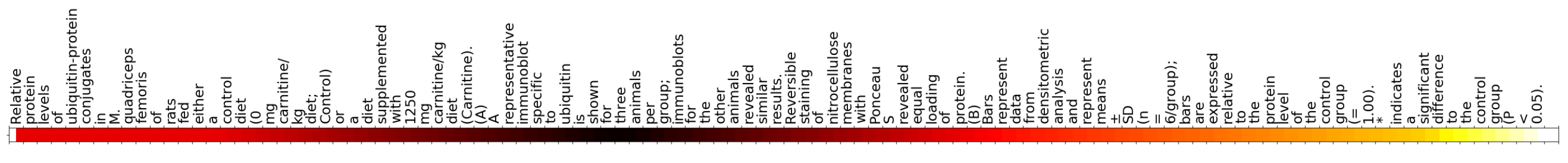}
    \caption{Sample caption activation heatmaps. Darker means higher activation.}
   \label{fig:textheat}
\end{figure*}

All the captions show a strong semantic correspondence with their associated figures. Figure~\ref{fig:textheat} shows the activation heatmaps for two sample captions, calculated on the embeddings layer of the language subnetwork. The upper one corresponds to the fourth column left-right and third figure top-down in figure~\ref{fig:textfeat}. Its caption reads: "The Aliev-Panfilov model with $\alpha=0.01$\ldots The phase portrait depicts trajectories for distinct initial values $\varphi_0$ and $r_0$\ldots". Below, (first column, fourth figure in figure~\ref{fig:textfeat}): "Relative protein levels of ubiquitin-protein conjugates in M. quadriceps\ldots A representative immunoblot specific to ubiquitin\ldots". Consistently with our analysis, activation focuses on the most relevant tokens for each text feature: "Aliev-Panfilov model" and "immunoblot", respectively.

\section{Conclusions} 
\label{sec:con}
There is a wealth of knowledge in scientific literature and only a fraction of it is text. However, understanding scientific figures is a challenging task for machines, which is beyond their ability to process natural images. In this paper, we provide empirical evidence of this and show that co-training text and visual features from a large corpus of scientific figures and their captions in a correspondence task (FCC) is an effective, flexible and elegant unsupervised means towards overcoming such complexity. We show how such features can be significantly improved by enriching them with additional knowledge sources and, particularly, structured KGs. We prove the benefits of our approach against supervised baselines and in different transfer learning tasks, including text and visual classification and multi-modal machine comprehension applied to question answering, with results generally beyond the state of the art. In the future, it will be interesting to further the study of the interplay between the semantic concepts explicitly represented in different KGs, contextualized  embeddings e.g. from SciBERT~\cite{Beltagy2019SciBERT}, and the text and visual features learnt in the FCC task. We also plan to continue to charter the knowledge captured in such features and to pursue the optimization and practical application of our approach.

\section*{Acknowledgments}
The research reported in this paper is supported by the EU Horizon 2020 programme, under grants European Language Grid-825627 and Co-inform-770302.

\bibliographystyle{named}
\bibliography{look_read_enrich}

\end{document}